\documentclass[10pt,conference]{IEEEtran}

\usepackage{url}
\usepackage{xcolor}
\usepackage{color}
\usepackage{graphicx}
\usepackage{amssymb}
\usepackage{amsmath,amsthm,paralist}
\usepackage{amsbsy}
\usepackage{cite}
\usepackage{algorithm,algorithmic}
\usepackage{afterpage}
\usepackage{multirow}
\usepackage{caption}
\usepackage{bbm}

\usepackage{epstopdf}

\usepackage{comment}

\usepackage{hyperref}
\hypersetup{
    colorlinks=true, 
    linktoc=all,     
    linkcolor=blue,  
}

\allowdisplaybreaks

\theoremstyle{remark}

\newcommand{\bitem}{\begin{itemize}}
\newcommand{\eitem}{\end{itemize}}

\newcommand{\supp}{\mathrm{supp}}
\newcommand{\beqn}{\begin{equation}}
\newcommand{\eeqn}{\end{equation}}
\newcommand{\balign}{\begin{align}}
\newcommand{\ealign}{\end{align}}

\def \R {\mathbb{R}}

\begin{document}

\title{An Asynchronous Parallel Approach to Sparse Recovery}

\author{\authorblockN{Deanna Needell}
\authorblockA{Department of Mathematical Sciences\\
Claremont McKenna College\\
Claremont, CA 91711, USA \\
Email: dneedell@cmc.edu}
\and
\authorblockN{Tina Woolf}
\authorblockA{Institute of Mathematical Sciences\\
Claremont Graduate University\\
Claremont, CA 91711, USA\\
Email: tina.woolf@cgu.edu}}

\maketitle

\begin{abstract}
Asynchronous parallel computing and sparse recovery are two areas that have received recent interest. Asynchronous algorithms are often studied to solve optimization problems where the cost function takes the form $\sum_{i=1}^M f_i(x)$, with a common assumption that each $f_i$ is \textit{sparse}; that is, each $f_i$ acts only on a small number of components of $x\in\R^n$. Sparse recovery problems, such as compressed sensing, can be formulated as optimization problems, however, the cost functions $f_i$ are \textit{dense} with respect to the components of $x$, and instead the signal $x$ is assumed to be sparse, meaning that it has only $s$ non-zeros where $s\ll n$. Here we address how one may use an asynchronous parallel architecture when the cost functions $f_i$ are not sparse in $x$, but rather the signal $x$ is sparse. We propose an asynchronous parallel approach to sparse recovery via a stochastic greedy algorithm, where multiple processors asynchronously update a vector in shared memory containing information on the estimated signal \textit{support}. We include numerical simulations that illustrate the potential benefits of our proposed asynchronous method. 
\end{abstract}

\section{Introduction}

Technological advances in data gathering systems have led to the rapid growth of big data in diverse applications. At the same time, the recent emergence of inexpensive multicore processors, with the number of cores on each workstation on the rise, has motivated the study of parallel computing strategies. This has presented a challenge to many existing and popular algorithms that are designed to run iteratively and sequentially. 

One possible approach to this problem is \textit{synchronous} parallel computing, which assigns tasks to multiple cores and then waits for all cores to complete before the next step begins. Of course, the drawback of this approach is that all cores must wait for the slowest core to finish, even if the remaining cores all complete their computation quickly. An alternative approach, and one that has received much recent interest, is \textit{asynchronous} parallel computing. In an asynchronous system, all cores run continuously, thus eliminating the idle time present in the synchronous approach, and all cores have access to shared memory and are able to make updates as needed.  

Asynchronous algorithms are often studied to solve optimization problems where the cost function takes the form $\sum_{i=1}^M f_i(x)$. The decision variable $x\in\R^n$ is updated iteratively, and its current state is accessible in shared memory by all processors. Many approaches to asynchronous parallel computing for solving optimizations problems of this form assume that each $f_i$ is sparse, which means that each $f_i$ acts only on a small number of components of $x$; therefore, this implies that individual core computations only depend on and update a small number of coordinates of $x$ (e.g., \cite{RechtRWN_hogwild11,DuchiJM_Estimation}). The benefit of this setup is that memory overwrites are rare, making it unlikely for the progress of faster cores to be overwritten by updates from slower cores.

There has also been a surge of interest in sparse recovery problems; for instance, literature in \textit{compressed sensing} (e.g., \cite{Donoh_Compressed,CandeRT_Stable,CandeRT_Robust}) addresses the problem of recovering a sparse vector $x\in\R^n$ from few nonadaptive, linear, and possibly noisy measurements of the form $y=Ax+z$, where $A\in\R^{m\times n}$ is the measurement matrix and $z\in\R^m$ is noise. 
Recovering $x$ from the noisy measurements $y$ can be formulated as the optimization problem, 
\begin{align}\label{compressed sensing}
\min_{\tilde{x}\in\R^n} \frac{1}{2m}\|y-A\tilde{x} \|_2^2 \quad \mbox{ subject to } \quad \|\tilde{x} \|_0 \leq s.
\end{align}
It is then natural to ask whether we can apply asynchronous parallel computing to solve (\ref{compressed sensing}). The challenge, however, is that the cost function depends on $A$, which is typically not taken to be sparse (e.g., $A$ is commonly taken to have standard i.i.d. Gaussian entries). This is in stark contrast to the typical assumptions made in the asynchronous parallel computing literature since the cost function is \textit{dense} in the components of $x$, and instead the signal $x$ is assumed to be sparse. Indeed, since the signal $x$ is sparse, it is very likely that the same non-zero entries will be updated from one iteration to the next, while the remaining entries are set to zero to maintain a sparse solution. If executed asynchronously, then memory overwrites would be frequent, and a slow core could easily ``undo" the progress of previous updates by faster cores. 

{\bfseries Contribution:} In this paper, we consider one of the stochastic greedy algorithms studied in \cite{NguyenNW_stochastic14} for sparse recovery. Focusing on the compressed sensing problem, we propose a strategy for utilizing the algorithm asynchronously despite the matrix $A$ (and thus the cost function) not being sparse. Instead of having the current solution estimate in shared memory, a current estimate of the location of the non-zeros of the signal will be shared and available to each core. Thus, we provide a solution that merges asynchronous parallel architectures with sparse recovery problems that have iterative updates which are not sparse, and will be a springboard to other more general approaches.

\section{Relation to Prior Work}

The seminal text \cite{Bertsekas_parallel} provides foundational work for parallel and distributed optimization algorithms. More recently, \cite{RechtRWN_hogwild11} studies an asynchronous variant of stochastic gradient descent called \textsc{Hogwild}!. A key assumption in their analysis is that the cost function of the optimization problem is sparse with respect to the decision variable, meaning that most gradient updates only modify small and distinct parts of the solution. Much of the recent literature on asynchronous parallel computing borrows from the framework proposed in \cite{RechtRWN_hogwild11}. \cite{DuchiJM_Estimation} also studies stochastic optimization when the cost function is sparse, and proposes two asynchronous algorithms with their analysis influenced by that in \cite{RechtRWN_hogwild11}. Also following the technique in \cite{RechtRWN_hogwild11}, \cite{LiuWS_AKacz} presents an asynchronous parallel variant of the randomized Kaczmarz algorithm for solving the linear system $Ax=y$ when $A$ is large and sparse. \cite{AvronDG_Revisiting} is interested in the same linear system with $A$ symmetric positive definite and presents an asynchronous solver; note that the convergence rate analysis again depends on the sparsity of the matrix.  Asynchronous parallel stochastic coordinate descent algorithms (which are clearly not designed for sparse solutions) are proposed in \cite{LiuWRBS__AsyncSCD} and \cite{LiuW_AsyncSCD}, which also follow the model of \cite{RechtRWN_hogwild11}. Other recent and relevant work on asynchronous parallel algorithms and analysis frameworks includes \cite{DuchiCR_AsyncConvex,LianHLL_AsyncSGD,HuoH_AsyncVariance,ZhaoL_Fast,FeyzmahdavianAJ_Mini,ManiaPPRRJ_Perturbed,CannelliFKS_AsyncBigData,PengXYY_ARock,LeblondPL_Asaga,DefazioBL_Saga,Davis_PALM,DavisEU_StochasticAPALM}.

In the sparse recovery literature, popular greedy recovery algorithms include IHT \cite{BlumeD_Iterative}, OMP \cite{TroppG_Signal}, and CoSaMP  \cite{NeedeT_CoSaMP}. Most relevant to our work is IHT, which we briefly review here. Starting with an initial estimation $x^1 = 0$, the IHT algorithm computes the following recursive update,
\begin{align}
x^{t+1} = \mathcal{H}_s(x^t + A^\star(y-Ax^t)),
\end{align}
where $\mathcal{H}_s(a)$ is the thresholding operator that sets all but the largest (in magnitude) $s$ coefficients of $a$ to zero. The work \cite{NguyenNW_stochastic14} proposes a stochastic variant of IHT called StoIHT, which is the algorithm of focus here. Note that \cite{NguyenNW_stochastic14} also proposes a stochastic variant of GradMP \cite{NCT_gradMP_2013_J} which is based on CoSaMP; our work here can easily be generalized to these other algorithms as well.

\section{Asynchronous Sparse Recovery with Tally Updates}

\textbf{Notation:} We denote by $[M]$ the set $\{1,2,\dots,M\}$ and let $p(1),\dots,p(M)$ be the probability distribution of an index $i$ selected at random from the set $[M]$, so that $\sum_{i=1}^M p(i) = 1$. Let $A^\star$ denote the conjugate transpose of the matrix $A$. For a vector $a\in\R^n$, let $\supp_s(a)$ be the operator that returns the set of cardinality $s$ containing the indices of the largest (in magnitude) elements of $a$. For a set $\Gamma$, let $a_{\Gamma}$ denote the vector $a$ with everything except the components indexed in $\Gamma\subset\{1,\dots,n\}$ set to zero.

As described above, suppose we observe $y=Ax+z$, where $x$ is $s$-sparse and supported on $T\subset\{1,\dots,n\}$ (that is, $\|x\|_0 = |\{i:x_i\neq 0 \}| = |T| \leq s \ll n$). To recover $x$ from the noisy measurements $y$, we aim to solve the optimization problem (\ref{compressed sensing}).
Note that we can also express the cost function in (\ref{compressed sensing}) as 
$$\frac{1}{2m}\|y-A\tilde{x} \|_2^2 = \frac{1}{M}\sum_{i=1}^M \frac{1}{2b}\|y_{b_i} - A_{b_i}\tilde{x} \|_2^2,$$ 
where $y$ has been decomposed into non-overlapping vectors $y_{b_i}$ of size $b$, $A_{b_i}$ has been decomposed into non-overlapping $b\times n$ submatrices of $A$, and $M=m/b$ (which for simplicity we assume is integral). Notice that the cost function now takes the form $\sum_{i=1}^M f_i(x)$, and each $f_i(x)= \frac{1}{2bM}\|y_{b_i} - A_{b_i}x \|_2^2$ accounts for a block of observations of size $b$. The StoIHT algorithm from \cite{NguyenNW_stochastic14} for solving (\ref{compressed sensing}), specialized to the compressed sensing setting, is shown in Algorithm \ref{alg:StoIHT}, where $\gamma$ denotes a step-size parameter. The recovery error of the algorithm depends on the block size $b$; we refer the reader to \cite{NguyenNW_stochastic14} for details.

\begin{algorithm}[ht]
\caption{StoIHT Algorithm \cite{NguyenNW_stochastic14}} 
\label{alg:StoIHT}
\begin{algorithmic}
\STATE \textbf{input:} $s$, $\gamma$, $p(i)$, and stopping criterion
\STATE \textbf{initialize:} $x^1$ and $t=1$ 
\REPEAT
\STATE
\begin{tabular}{ll}
\textbf{randomize:} & select $i_t\in[M]$ with probability $p(i_t)$ \\
\textbf{proxy:} & $b^t = x^t + \frac{\gamma}{Mp(i_t)}  A^\star_{b_{i_t}}(y_{b_{i_t}}-A_{b_{i_t}}x^t)$ \\
\textbf{identify:} & $\Gamma^t = \supp_s (b^t)$ \\
\textbf{estimate:} & $x^{t+1} = b^t_{\Gamma^t}$ \\
\ & $t = t+1$	
\end{tabular}
\UNTIL{halting criterion \textit{true}}
\STATE \textbf{output:} $\hat{x} = x^t$
\end{algorithmic}
\end{algorithm}

In the asynchronous approach, each core will execute its own slightly modified version of Algorithm \ref{alg:StoIHT}. In order to avoid having each core update the entire solution in shared memory at each iteration, we propose to instead keep a \textit{tally} vector $\phi\in\R^n$ in shared memory. The tally $\phi$ will contain information on the estimated support locations identified by each core's most recent iteration. 

The steps executed by \textit{each} core at each iteration are detailed in Algorithm \ref{alg:StoIHT tally}, where the tally vector $\phi$ is available for read and write by each core. Note that in Algorithm \ref{alg:StoIHT tally} the iteration number $t$ and the iterate $x^t$ are local to each core. Also note that in the tally update step, the tally is incremented by $t$ (the core's local iteration number) on $\Gamma^t$ and decremented by $t-1$ on $\Gamma^{t-1}$ (as mentioned in \cite{RechtRWN_hogwild11}, these operations can be performed atomically on most modern processors). This is to provide more weight to faster cores that are further along in the algorithm and should thus be able to provide a better support estimate, and less weight to slower cores. In order to maintain only the information from each core's most recent iteration, the tally contribution from the previous iteration $t-1$ is removed. 

It is worth mentioning that \textit{inconsistent} reads, where components of the shared memory vector variables may be written by some cores while being simultaneously read by others, is discussed in the current literature on asynchronous computing. It is desirable to incorporate this feature into any models and analyses to more faithfully represent modern computational architectures. Our approach is not immune to inconsistent reads of the tally $\phi$ by any means, however, the hope is that the algorithm will be more robust to inconsistent reads of $\phi$ than a current solution estimate since its use in the algorithm is more passive, and that an inconsistently read version of $\phi$ will still provide valuable information on the support locations of the signal. Although not addressed here, analyses and simulations of this impact are certainly of interest. 

\begin{algorithm}[ht]
\caption{Asynchronous StoIHT Iteration} 
\label{alg:StoIHT tally}
\begin{algorithmic}
\STATE Each core performs the following at each iteration. The tally vector $\phi$ is available to each core.\\
\STATE
\begin{tabular}{ll}
\textbf{randomize:} & select $i_t\in[M]$ with probability $p(i_t)$ \\
\textbf{proxy:} & $b^t = x^t + \frac{\gamma}{Mp(i_t)}  A^\star_{b_{i_t}}(y_{b_{i_t}}-A_{b_{i_t}}x^t)$ \\
\textbf{identify:} & $\Gamma^t = \supp_s (b^t)$\\
& $\widetilde{T}^t = \supp_s (\phi)$ \\
\textbf{estimate:} & $x^{t+1} = b^t_{\Gamma^t\cup\widetilde{T}^t}$ \\
\textbf{update tally:} & $ \phi_{\Gamma^t} = \phi_{\Gamma^t} + t$ \\
& $\phi_{\Gamma^{t-1}} = \phi_{\Gamma^{t-1}} - (t-1)$ \\
\ & $t = t+1$	
\end{tabular}
\end{algorithmic}
\end{algorithm}

\section{Simulations}

In this section, we provide encouraging experimental results for the proposed asynchronous tally update scheme. In all of the experiments, we take the signal dimension $n=1000$, the sparsity level $s=20$, the number of measurements $m=300$, the block size $b=15$, the step-size $\gamma=1$, and the initial estimate $x^1=0$. The algorithms exit once $\|y-Ax^t \|_2$ drops below the tolerance $10^{-7}$ or a maximum of 1500 iterations is reached.

\subsection{StoIHT with an Accurate Support Estimate}

\begin{figure}[!htbp]
\centering

\includegraphics[height=1.8in]{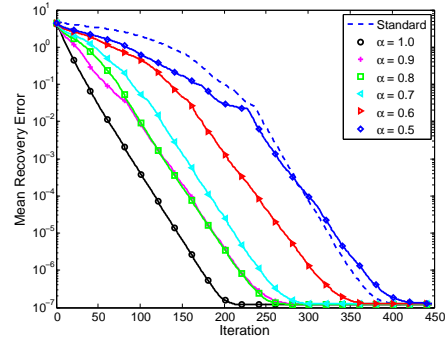} \\
\caption{Mean recovery error over 50 trials versus iteration of StoIHT and a modified version of StoIHT.
In the modified StoIHT, we execute Algorithm \ref{alg:StoIHT} where the estimation step is performed by projecting $b^t$ onto $\Gamma^t\cup\widetilde{T}$ at each iteration, where $\widetilde{T}$ has accuracy $\frac{|\widetilde{T}\cap T|}{|\widetilde{T}|}=\alpha$.}
\label{fig::support estimate}
\end{figure}

Our first experiment provides evidence that performing the estimation step onto the top $s$ coefficients in addition to a set that accurately describes the true signal support will increase the convergence rate of the standard StoIHT algorithm. That is, we execute Algorithm \ref{alg:StoIHT} with the modification that $x^{t+1} = b^t_{\Gamma^t\cup\widetilde{T}}$, where $\widetilde{T}$ estimates the true support $T$ with $|\widetilde{T}|=s$ and accuracy $\frac{|\widetilde{T}\cap T|}{|\widetilde{T}|} = \alpha$. 
Figure \ref{fig::support estimate} compares the mean recovery error as a function of iteration over 50 trials of the standard StoIHT algorithm (Algorithm \ref{alg:StoIHT}) with the modified StoIHT algorithm as described for various values of $\alpha$. 
Indeed, for $\alpha>0.5$, fewer iterations are needed on average for the algorithm to converge. When $\alpha=1$, on average roughly half as many iterations as the standard algorithm are needed for convergence. We emphasize that this experiment, as a proof of concept, has no parallel implementation, and hence iterations are comparable to runtime. This result suggests that the asynchronous approach using Algorithm \ref{alg:StoIHT tally} could lead to speedups as long as the tally $\phi$ becomes accurate fast enough.

\begin{figure}[!htbp]
\centering
\includegraphics[height=2.1in]{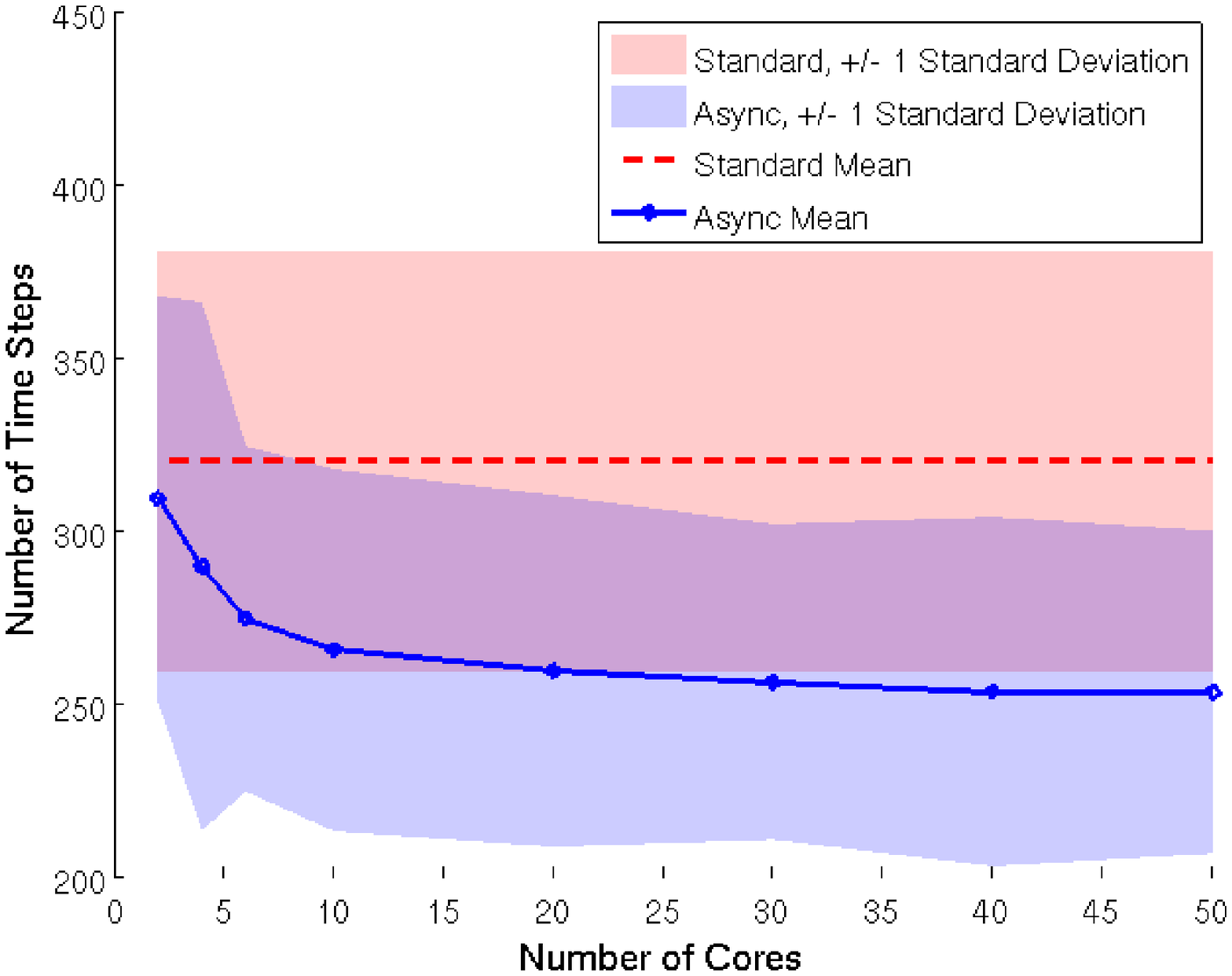} \\
\includegraphics[height=2.1in]{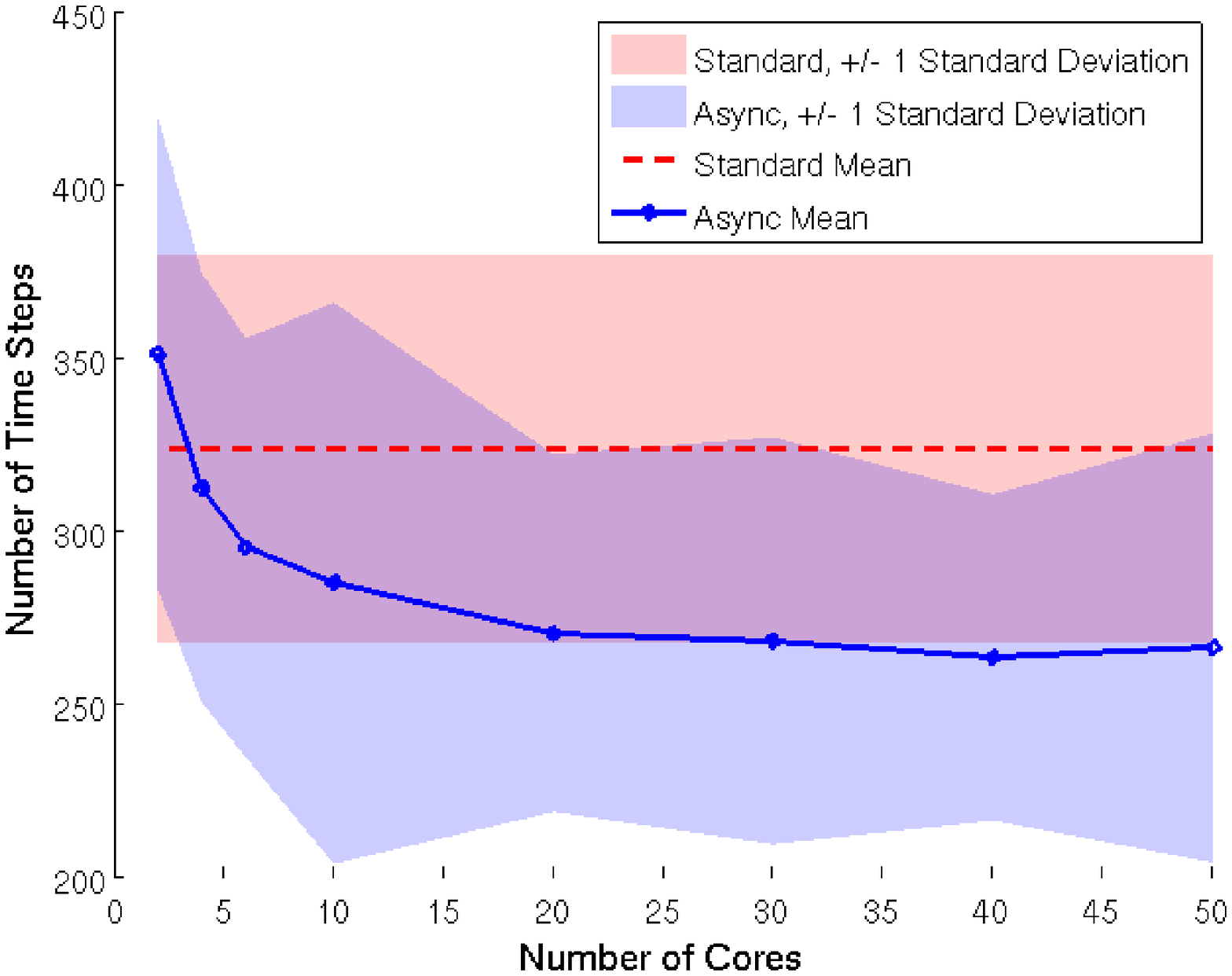} \\
\caption{Comparison of the number of time steps executed until convergence versus the number of cores used in the asynchronous StoIHT method. The heavy line denotes the mean number of time steps over 500 trials, and the boundaries of the shaded region indicate $\pm1$ standard deviation from the mean.
(Upper) All cores are simulated to complete an iteration in a single time step; (lower) half of the cores are ``slow" and complete an iteration only once out of every four time steps.}
\label{fig::tally}
\end{figure}

\subsection{Asynchronous StoIHT}

Here, we simulate the execution of asynchronous StoIHT with $c$ cores, where each core uses the iteration defined in Algorithm \ref{alg:StoIHT tally}. For clarity, define a \textit{time step} to be the amount of time needed for the fastest core to complete an iteration of Algorithm \ref{alg:StoIHT tally}. First, we assume each core takes the same amount of time to perform an iteration; thus, in a single time step, all $c$ cores complete an iteration of Algorithm \ref{alg:StoIHT tally}. We also assume that an iteration of Algorithm \ref{alg:StoIHT} takes a single time step. 
When executing the $t$-th time step, and hence the $t$-th iteration for each core, every core utilizes the same set $\widetilde{T}^t$ identified by the tally $\phi$. Once each core completes its estimation step, the tally is updated via $\Gamma^t$ from \textit{each} core. As soon as \textit{any} core achieves the exit criteria at its local iteration $t$, the algorithm terminates, and $t$ time steps are recorded as the number of time steps until exit is achieved. The upper plot of Figure \ref{fig::tally} displays the mean number of time steps until exit, over 500 trials, for both the standard and asynchronous StoIHT algorithms, plotted against the number of cores used in the asynchronous method. The shaded regions indicate $\pm$1 standard deviation from the mean. Since the standard method does not depend on the number of cores, horizontal lines are shown.
 Notice that the mean number of time steps required in the asynchronous method is always less than the standard algorithm; therefore, a speedup in the total time for the algorithm to converge is expected when executed asynchronously. 

Next, we modify the previous experiment to simulate the impact of slow cores. We take half of the cores to be ``fast," meaning that they continue to update the shared tally at each time step; the other half of the cores, however, are ``slow" and only complete an iteration and update the tally once out of every four time steps. The lower plot of Figure \ref{fig::tally} displays the mean number of time steps until exit versus the number of cores in the asynchronous method. For $c=2$, no improvement is gained from the asynchronous method on average, however, improvement is observed for the larger values of $c$ tested.

\section{Conclusions and Future Work}

In this paper, we have proposed an asynchronous parallel variant of an existing stochastic greedy algorithm for sparse recovery. Our method is distinct from much of the existing literature on asynchronous algorithms because: 1) sparsity assumptions on the cost function, which are common to the existing literature, are not necessary, and 2) the current solution iterate is not available in shared memory, but instead a \textit{tally} vector containing the latest information from the cores on the estimated support of the signal is shared and utilized; this approach provides necessary robustness to asynchronous updates and inconsistent reads even when traditional updates cannot be made sparse.

A similar approach could also be applied to the second stochastic greedy algorithm studied in \cite{NguyenNW_stochastic14}, namely, StoGradMP. Although we specialized to the compressed sensing problem, both StoIHT and StoGradMP are studied for general sparse recovery problems in \cite{NguyenNW_stochastic14}; thus, it would be interesting to explore the proposed idea in other settings. It would also be beneficial to develop theory for the proposed approach, perhaps building from the theory in \cite{NguyenNW_stochastic14} and the current literature analyzing asynchronous algorithms, and include the incorporation of architecture realities such as inconsistent reads.

\section*{Acknowledgment}
The work of Deanna Needell was partially supported by NSF CAREER grant \#1348721 and the Alfred P. Sloan Foundation. The work of Tina Woolf was partially supported by NSF CAREER grant \#1348721.

\bibliographystyle{plain}
\bibliography{bib}

\end{document}